# The minimal computational substrate of fluid intelligence

Amy PK Nelson[1], Joe Mole[2,3], Guilherme Pombo[1], Robert J Gray[1], James K Ruffle[1], Edgar Chan[2,3], Geraint E Rees[,4], Lisa Cipolotti*[2,3], Parashkev Nachev*[1]

1. High Dimensional Neurology Group, UCL Queen Square Institute of Neurology, University College London, Russell Square House, Bloomsbury, London WC1B 5EH, UK
2. Department of Neuropsychology, National Hospital for Neurology and Neurosurgery, London, UK
3. UCL Queen Square Institute of Neurology, London, UK
4. Faculty of Life Sciences, University College London, Gower Street, London WC1E 6BT, UK

   * contributed equally

**Abstract**

The quantification of cognitive powers rests on identifying a behavioural task that depends on them. Such dependence cannot be assured, for the powers a task invokes cannot be experimentally controlled or constrained *a priori*, resulting in unknown vulnerability to failure of specificity and generalisability. Evaluating a compact version of Raven's Advanced Progressive Matrices (RAPM), a widely used clinical test of fluid intelligence, we show that LaMa, a self-supervised artificial neural network trained solely on the completion of partially masked images of natural environmental scenes, achieves human-level test scores *a prima vista,* without any task-specific inductive bias or training. Compared with cohorts of healthy and focally lesioned participants, LaMa exhibits human-like variation with item difficulty, and produces errors characteristic of right frontal lobe damage under degradation of its ability to integrate global spatial patterns. LaMa's narrow training and limited capacity—comparable to the nervous system of the fruit fly—suggest RAPM may be open to computationally simple solutions that need not necessarily invoke abstract reasoning.

**Introduction**

Cognitive tests inevitably rely on reductive operationalizations of the abilities they seek to measure. The broader the cognitive domain, the harder it is to design a behavioural task that captures its distinctive neural substrate. The difficulty is arguably greatest for fluid intelligence—the ability to solve novel problems without prior experience—where domain-specific constraint is definitionally lacking. Indeed, novelty and abstraction appear to be the only substantive manipulable task dimensions here maximised to leave no 'higher' ability meaningfully in play. The most widely used compact test of fluid intelligence in clinical settings, Raven's Advanced Progressive Matrices (RAPM)[1], employs masked structured patterns of abstract geometric shapes to be completed from a choice of options, assuming that success implies explicitly identifying and applying the underlying compositional rule. But this, and other assumptions common to such tests, are not secure: the plausibility of generalisation—and interpretation in mechanistic terms—requires corroboration beyond performance on the test itself. Crucially, the traditional set of corroborating 'validities'—construct validity, content validity, and criterion validity[2–4]—omits what we call 'substrate validity': determining the minimal computational substrate sufficient to perform the behavioural task successfully on comparable training. Yet if the minimal substrate of a task is substantially narrower in scope than the target ability, any observed association need not imply necessity, potentially limiting interpretation and generalisability.

The omission of substrate validity is explained by the formidable difficulty of creating artificial substrates capable of replicating human cognitive powers. A valid comparative substrate must be *both* uncontaminated by the target ability *in its design* and not explicitly *trained* on the specific task or anything like it. For example, we obviously cannot use an electronic calculator as a benchmark for the substrate of arithmetic skill, for the rules of arithmetic are embodied in the design. Equally, we cannot use a discriminative machine vision model trained and tested on faces as a benchmark for visual recognition, for the model may be easily overfitted to the task. Only minimally constrained, self-supervised model architectures are acceptable candidates: a form the revolution in deep learning has only recently made possible. Moreover, where—as here—task novelty is an essential element, *a prima vista*, 'zero-shot' operation without any prior exposure is required. But now that it is feasible, we suggest that substrate validity analysis ought to be considered in the design and evaluation of cognitive tests, especially those whose operationalization is challenging.

Here we demonstrate the implementation of this approach, and identify a minimal computational substrate capable of both *human-level* and *human-quality* performance on RAPM. Recognizing that solving RAPM-style problems minimally requires the interpolation of interacting spatial patterns, we hypothesise that image 'in-painting' models, trained solely on the completion of partially occluded unrelated natural images, may be sufficiently powerful to solve them. We examine a family of compact, self-supervised 'in-painting' models satisfying the validity criteria, and show that psychometric performance within the human distribution—both healthy participants and patients with right frontal lobe lesions, a population with known impaired RAPM performance[5]—is achieved by a model with only 51 million trainable parameters, and no training beyond natural image in-painting. We further show through model ablation studies— 'computational lesion-deficit mapping'—that long range integration of periodic visual information is critical to RAPM performance, and that disruption of this component leads to error patterns seen in patients with selective deficits in RAPM arising from right frontal damage. Our analysis suggests that RAPM is at least in part open to solutions reliant on flexible integration of long-range spatially organized patterns.

## Results

*The distribution of RAPM performance in health and focal brain injury*

We examined a standard, validated, 12-item version of Raven's Advanced Progressive Matrices test[6] compact enough to be administered in a clinical context. Evaluating together a previously reported (n=77)[5] and a new set of patients with right frontal focal brain lesions (n=12) exhibiting reduced RAPM performance (N=89), and a subset (n=114) of previously reported healthy control participants (N=315) with complete RAPM response data[7], revealed performance distributions consistent with the wider literature (Figure 1, Table 1). The mean of the distribution, both for RAPM and the National Adult Reading Test (NART)—a measure of premorbid optimal functioning—was slightly higher than the reference average (Table 1), in reflection of the constitution of our local patient population. These distributions replicate the known properties of RAPM[6–8], and provide an index of its variability in the context of the focal brain injury most likely to impair it.

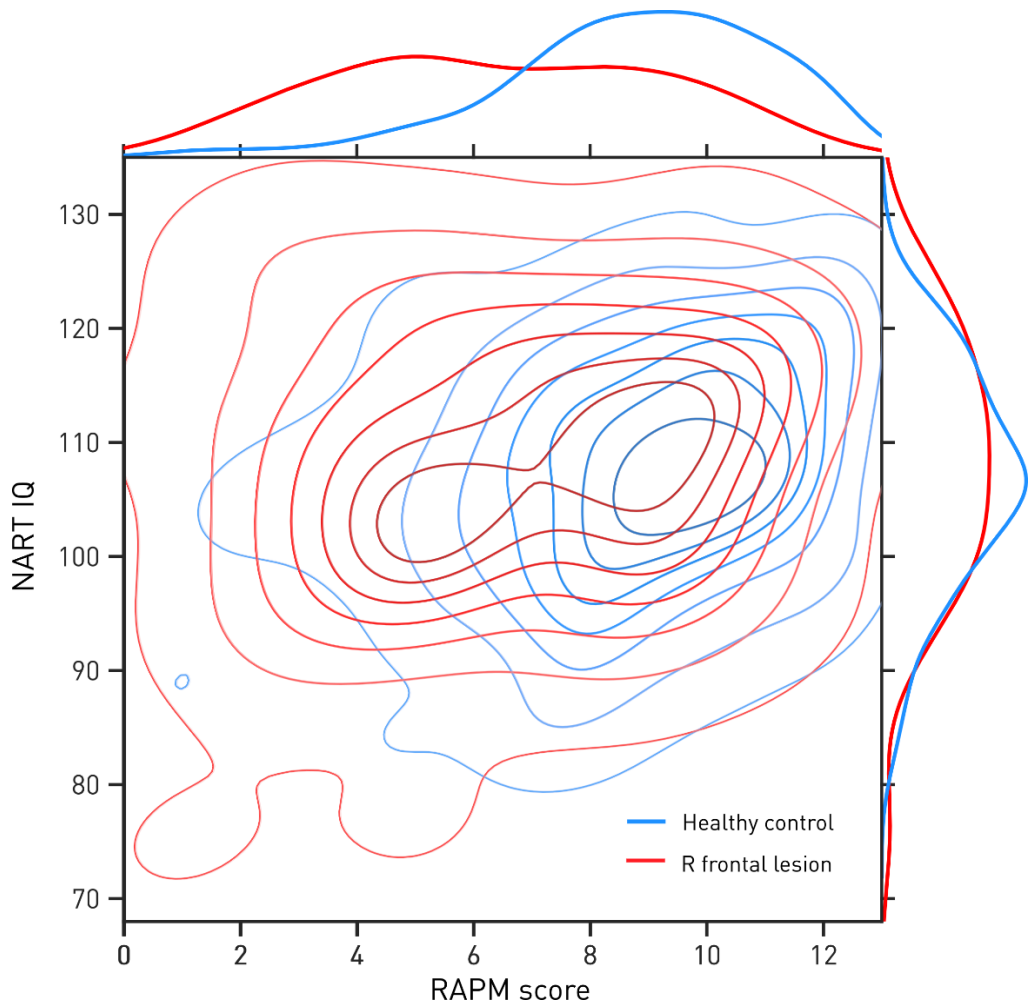

**Figure 1. The joint distributions of Raven's Advanced Progressive Matrices (RAPM) scores and National Adult Reading Test (NART) IQs in healthy controls (blue) and right frontal lesion patients (red), displayed as 2D kernel density estimates.** While the NART distributions are similar, the right frontal patients have a markedly left-shifted RAPM score distribution compared with healthy control participants (as previously reported[5]). Contour lines are iso-proportions of the density at 12.5% intervals. For visualisation purposes, the distributions are normalised independently for healthy control participants and right frontal patients.

**Table 1.** Population demographics and performance means (±1 standard deviation, or percent for categories).

|  | **Healthy Control** | **Right frontal lesion** |
|---|---|---|
| **N** | 315 | 89 |
| **Age** | 58.89 (39.41-78.37) | 49.50 (33.66-65.34) |
| **Sex (Male)** | 114 (36.19%) | 45 (50.56%) |
| **Years of education** | 14.26 (11.64-16.88) | 14.68 (10.84-18.51) |
| **RAPM** | 8.77 (6.48-11.06) | 6.42 (3.64-9.19) |
| **NART IQ** | 106.96 (97.32-116.59) | 108.04 (96.77-119.31) |

*A small 'in-painting' model can achieve human-level RAPM performance*

The completion of Raven's matrices may be efficiently conceived as requiring the interpolation of interacting spatial patterns. The difficulty is controlled by the number of dimensions on which the component figures interact and the complexity of the interaction, ranging from simple unidimensional progression along each plane, to additive or subtractive operations involving multiple figural dimensions: the cognitive demands—and underlying neural dependencies—may correspondingly differ[5]. Such 'spatial arithmetic' may arise naturally through superposition or occlusion of regularly structured objects: simple progression where two discretized spatial gradients unite, more complex addition or subtraction where non-linearly organized objects unite or occlude each other.

A minimal computational substrate could therefore be derived from deep generative models trained to complete diverse, partially occluded, natural images. Such models learn to 'in-paint' a designated region of an image from information contained in the remainder without prior exposure to images of exactly the same kind, interpolating such spatially organised patterns as they are able to discern from the training corpus. For a test of substrate validity to be licit, the model must not incorporate any prescribed test-specific inductive biases (e.g. explicit rules of spatial interpolation), cannot be trained with direct supervision on the test corpus or its variants, and must rely in training wholly on naturalistic images plausibly unconnected with the test corpus. An established family of 'in-painting' models—LaMa[9]—satisfies these requirements. LaMa is a simple, fully-convolutional, feed-forward, ResNet[10]-like artificial neural network with no substantive inductive bias beyond continuity of unspecified spatial patterns. It is trained solely on 4.5 million distinct images from the 'Places' dataset of in- and outdoor environmental scenes, with the sole objective of completing arbitrarily masked parts of an image. The dataset contains no psychometric tests of any kind, and such geometric patterns as occur are of the implicit kind contained in natural scenes.

To pose each question in a form answerable by an 'in-painting' model, we presented each test image with the region of the ninth cell of the 3x3 matrix marked for 'in-painting', and matched the in-painted result to the closest answer option by taking the mode of a panel of image similarity metrics. Though the human test does not require such explicit synthesis, this mechanism has the advantage of rendering the model's 'choice' legible to an external observer. LaMa, a comparatively small, 51 million parameter model, produced the correct response on 8 out of 12 questions (Figure 2A). By comparison, healthy human control participants scored a mean of 8.77 (+/- 1 SD = 6.48-11.06), placing LaMa well within 1 standard deviation of the healthy distribution. The RAPM score obtained by LaMa corresponded by Ordinary Least Squares coefficient to a healthy human NART IQ of 106[7]. The in-painted regions were clearly correct on visual inspection, across a wide variety of RAPM question types (Figure 3A&B).

*'In-painting' model performance depends on capacity to capture global spatial regularities*

To investigate the computational dependents of RAPM performance, we evaluated degraded variants of the LaMa model in a form of 'computational lesion-deficit mapping' (known as ablation study in computer science). We first evaluated a model with fewer parameters, trained on a smaller dataset (LaMa Small, 27M parameters[9]). Reasoning from the characteristic structure of RAPM-style problems that performance may depend on sensitivity to global periodic patterns augmented in LaMa by the use of Fast Fourier Convolutional (FFC) modules, we further evaluated a version that used conventional convolutions instead (LaMa FFC-ablated, 74M parameters[9]). LaMa Small achieved a score of 6, whereas the FFC-ablated model, despite the greater expressivity afforded by a larger number of trainable parameters, achieved a score of only 1 (Figure 2A). This effect was present not just on objective scoring, but also visually (Figure 3). While decreasing model and training set size had a moderate performance effect, ablating the model of FFC, the component that facilitates learning of periodic structure from the full receptive field of an image, diminished its performance to the level expected by chance.

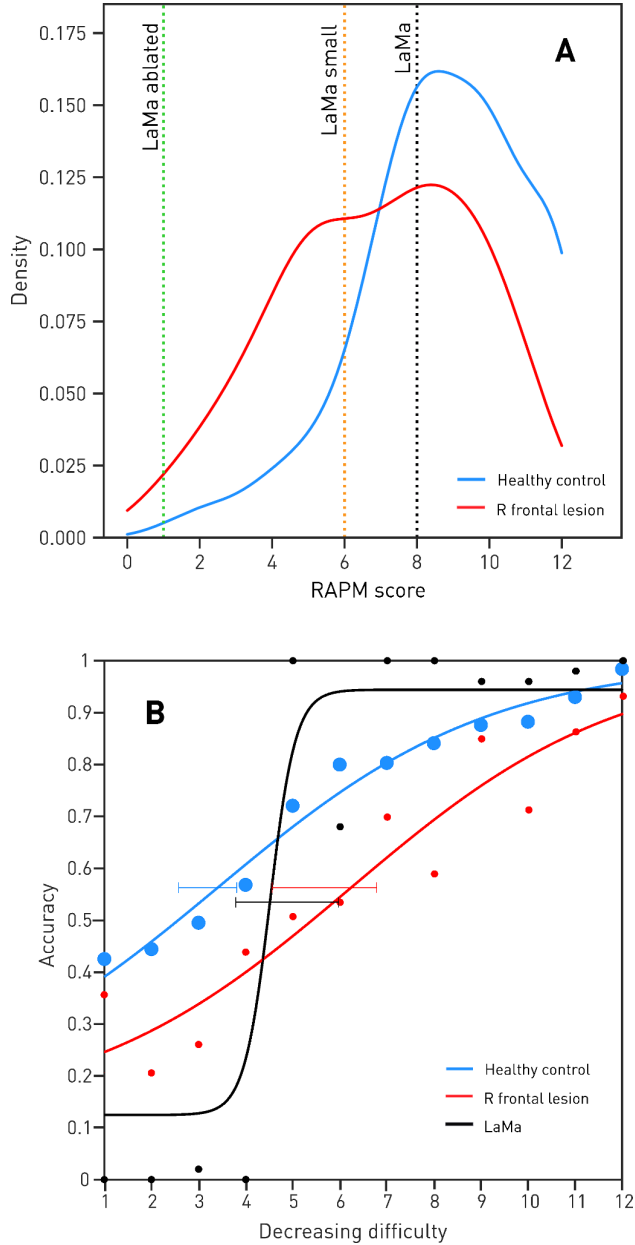

**Figure 2. The RAPM performance of LaMa models against the distributions of healthy controls and right frontal lesion patients.** (A) the distribution of healthy control (blue) and right frontal lesion patient (red) RAPM scores, displayed as kernel density estimates, and the performance of LaMa, LaMa FFC-ablated, and LaMa small models. LaMa, an image 'in-painting' model trained only on data from built and natural scenes, achieves a RAPM score well within the human distribution. While reducing model size results in a moderate reduction of RAPM score, a model ablated of FFC scores at chance. (B) the psychometric functions of healthy control (blue), right frontal lesion patients (red) and LaMa. The threshold, measured at 50% correct performance, was credibly different in right frontal lesion patients relative to healthy controls (Figure 3, 95% Bayesian credibility interval 4.56-6.78 item rank vs 2.56-3.81); the threshold of LaMa was within healthy control and right frontal lesion credibility intervals.

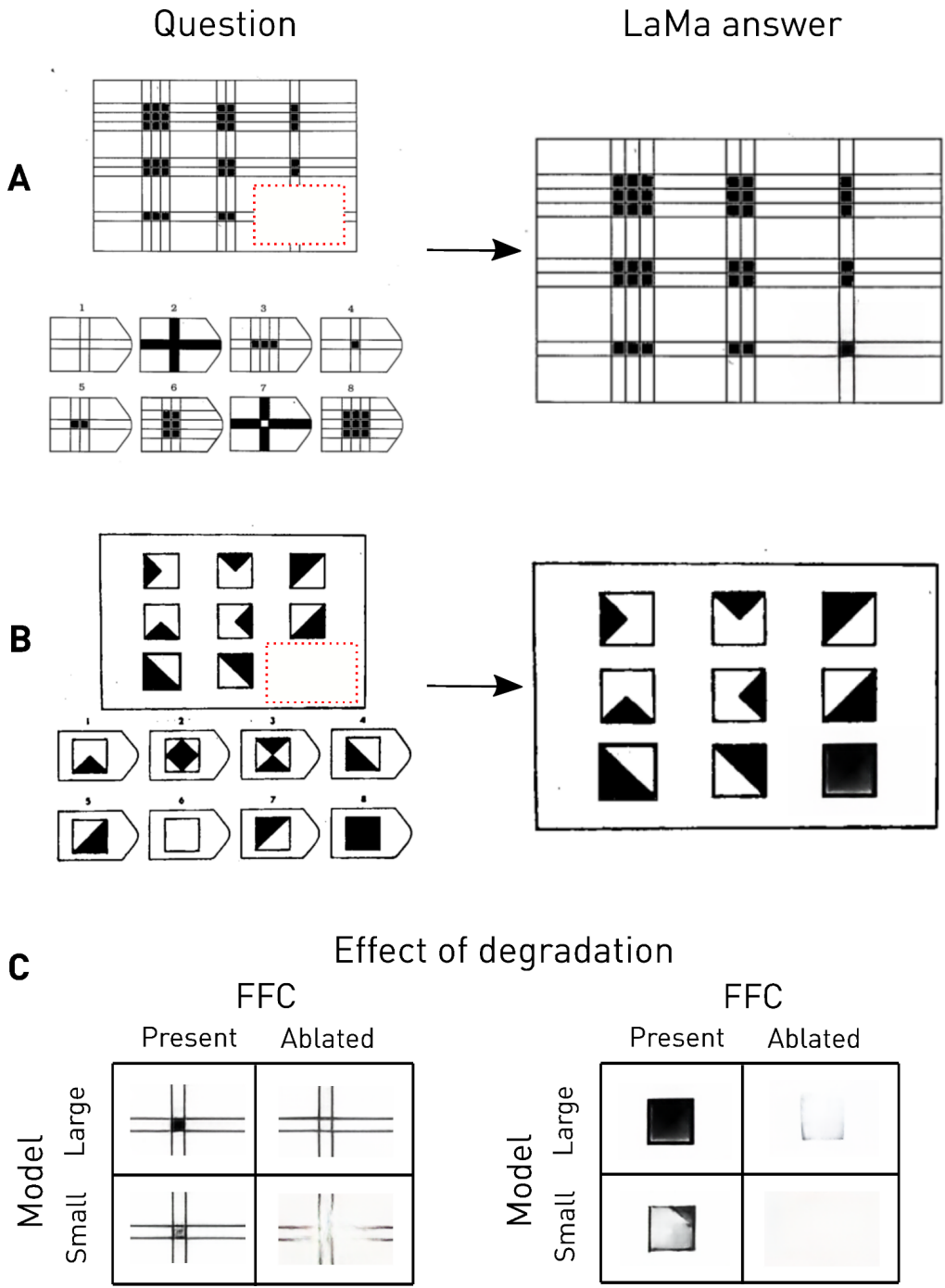

**Figure 3. Visual results of model 'in-painting' on sample RAPM questions.** (A, B) LaMa correctly in-paints the missing 'answer' region of RAPM questions. (C) Ablating the model of FFC modules results in failure to capture long-range visual patterns, but short-range patterns are propagated. Reducing model size results in less degraded performance. Models both reduced in size and ablated of FFC modules are incapable of completing the task at any level of difficulty. Note that for copyright reasons we are only able to show one real RAPM question (A); the second (B) merely resembles a real question. The quantitative analysis is confined to real RAPM questions only.

*Psychometric analysis*

A psychometric test may be described not just by its summary score, but by a function relating the probability of a correct response to the difficulty of the component item[11]. Quantifying performance within a Bayesian framework as a function of item difficulty derived from the healthy human cohort yielded well-fitting group-level psychometric functions across all groups (Figure 2B). LaMa psychometric functions were obtained by evaluating performance on 50 randomly modified versions of each RAPM image—varying brightness and the addition of Gaussian noise. The threshold of right frontal lesion patients, measured at 50% correct performance, credibly differed from healthy controls (Figure 2B, 95% Bayesian credibility interval 4.56-6.78 item rank vs 2.56-3.81 item rank), in line with our previous analysis[5]. LaMa's threshold interval—3.79 to 5.97—was comfortably within the 95% credibility intervals of both healthy control and right frontal lesion groups, and was obtained from a good fit, indicating that the model behaved broadly as a human psychometric subject. Note that the slope here, while steeper for the LaMa model, cannot be directly compared with human group psychometric function slopes since the former is derived from the function of a single 'agent' under image modifications to the input, whereas the latter summarises functions across a population. Bayesian methods are used here to facilitate inference to the absence of significant differences between groups.

*Analysis of errors*

Commission errors in Raven's matrix tasks are known to be non-random, reflecting consistent mistakes such as repetitions of neighbouring shapes, and incorrect or incomplete rule identification[5,7,12,13]. The errors of right frontal patients differed from those of control participants (Figure 4). Five right frontal error responses differed significantly from healthy controls: question 5 response 4 ($p<0.001$), question 6 response 8 ($p<0.001$), question 7 response 8 ($p=0.0011$), question 8 response 8 ($p<0.0001$), and question 10 response 4 ($p=0.0012$), corresponding to repetition, repetition, incorrect rule, incorrect rule, and repetition error-types respectively.

A comparative analysis of the LaMa models across all participants (healthy controls and patients with frontal or posterior lesions) showed significant correlation between participants who made any of the LaMa FFC-ablated model errors and the presence of right frontal lesions ($p=0.0018$) vs participants who did not make these errors (Table 2). There were no significant correlations between LaMa or LaMa small model errors and any other patient characteristics (Table 2). All observed values for each cell of the chi-squared contingency table were above 5.

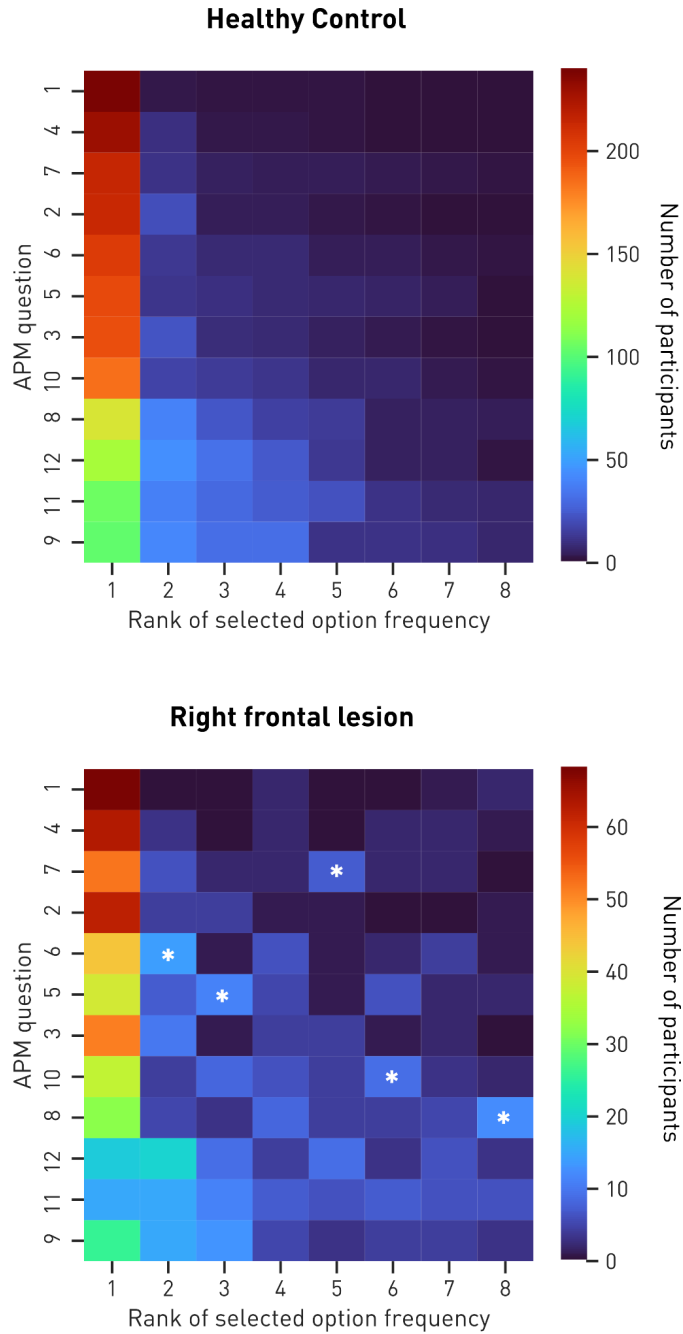

**Figure 4. RAPM response patterns in healthy controls and patients with right frontal lesions.** Responses are here represented on a grid where individual items are ordered by decreasing success rate (top to bottom) and decreasing selection frequency (left to right), both derived from the healthy control population. Each cell shows the corresponding response count across the healthy (A) and right frontal (B) cohorts, indexed by the colourmap. The majority response for each question is the correct one (first column). Note that right frontal patients differ not only in their overall scores but also their error patterns, significantly so for the question/response combinations indicated with a white asterisk. See main text for statistical details.

**Table 2.** Group characteristics of participants that make model-type errors vs those that do not. An asterisk identifies adjusted p-value <0.05 alpha. Categorical variables are counts (with fraction of total); continuous variables are mean (with 1 standard deviation).

|  | LaMa errors | No LaMa errors | LaMa FFC-ablated errors | No LaMa FFC-ablated errors | LaMa Small errors | No LaMa Small errors |
|---|---|---|---|---|---|---|
| **N** | 41 | 361 | 40 | 362 | 35 | 367 |
| **Age** | 51.2 (18.7) | 51.2 (18.7) | 58.5 (18.8) | 50.4 (18.5) | 57.4 (17.0) | 50.6 (18.8) |
| **Sex (Male)** | 17 (41.5%) | 159 (44.0%) | 23 (57.5%) | 153 (42.3%) | 19 (54.3%) | 157 (42.8%) |
| **Right frontal lesion** | 9 (22.0%) | 64 (17.7%) | 15 (37.5%) * | 58 (16.0%) | 9 (25.7%) | 64 (17.4%) |
| **Education years** | 14.4 (2.7) | 14.3 (3.0) | 14.3 (4.2) | 14.3 (2.9) | 13.5 (2.9) | 14.4 (3.0) |
| **NART** | 107.5 (9.6) | 107.2 (10.3) | 108.5 (10.8) | 107.1 (10.2) | 104.0 (12.5) | 107.6 (10.0) |

*Graph lesion-deficit mapping of FFC-ablation type errors*

Stochastic block modelling-based lesion-deficit mapping[14,5] (see Methods) revealed two discrete anatomical communities in the brain significantly associated with 'FFC-ablated error' types across patients with any lesion (frontal or posterior), controlling for pathologically-induced collateral patterns of damage: one in the right frontal pole (MNI x=8, y=66, z=3; posterior mean of intra-community edges: 'FFC-ablated error' 0.33, 95% credible interval (CI) 0.28-0.39 vs 'lesion co-occurrence' 0.23, 95% CI 0.19 to 0.26), the second at the intersection between the right middle frontal gyrus, inferior frontal gyrus, and precentral gyrus (MNI x=51, y=12, z=31; posterior mean of intra-community edges: 'FFC-ablated error' 0.24, 95% CI 0.20-0.28 vs 'lesion co-occurrence' 0.17, 95% CI 0.14 to 0.19) (Figure 5). This localisation should be cautiously interpreted given the marked class imbalance in the data (14 vs 94).

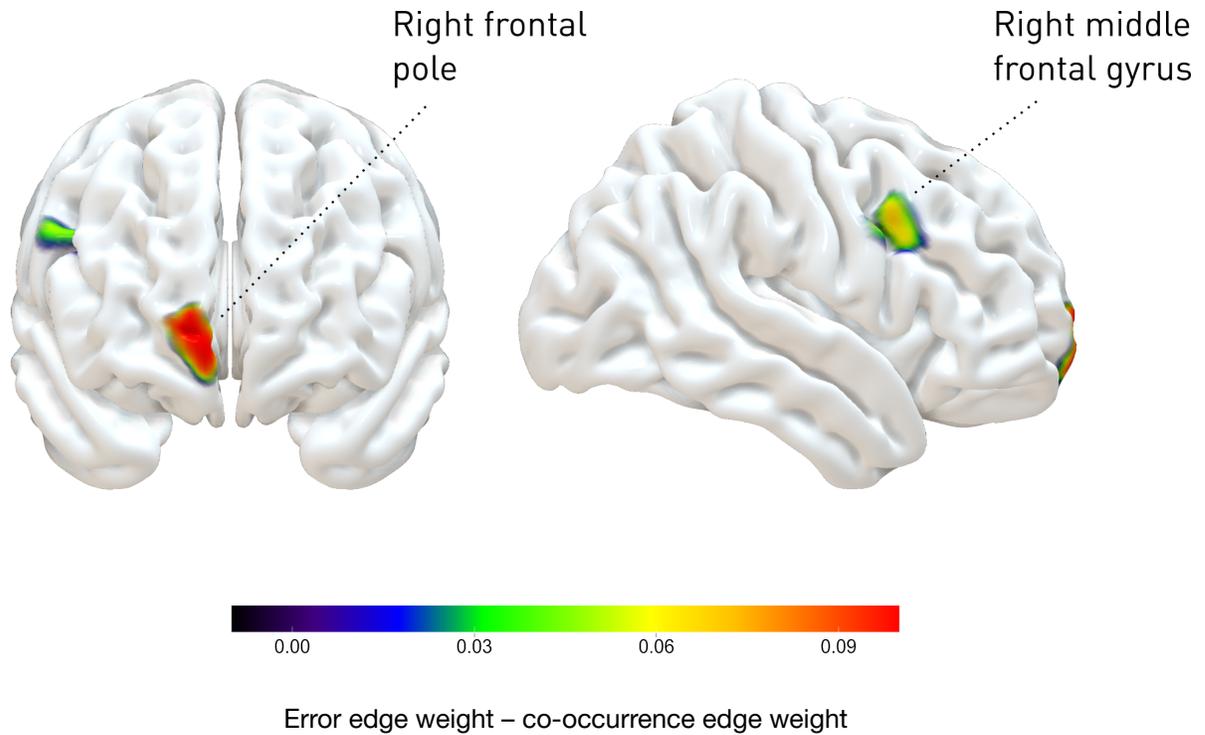

**Figure 5. Graph lesion-deficit mapping of FFC-ablation type errors in 106 patients with focal lesions distributed across the brain.** A layered stochastic block model of lesion data at 4x4x4mm resolution incorporating lesion co-occurrence and FFC-ablation type errors as edge covariates reveals two right frontal areas above the 95% credibility interval associated with errors. The colourmap shows the difference between the two edge weights, indicating a stronger association with FFC-ablation error type than with coincident patterns of lesion co-occurrence in the highlighted areas. See main text for statistical details. Note that with the comparatively low number of patients and in the presence of marked class imbalance, these localisations should be interpreted only as suggestive.

*The minimum computational substrate of RAPM*

We have seen that LaMa's powers are constitutionally narrowed to those necessary for the task of 'in-painting' natural scenes. The computational capacity is further limited by the number of trainable parameters, which allows comparison with the equivalent neural capacity of a range of biological organisms: the roundworm[15], fruit fly[16], honey bee[17], house mouse[18], cat[19] and human[20,21]. In keeping with established practice, we assumed approximate parity between the number of trainable model parameters and the number of neural synapses[22,23]. The closest neural equivalent of LaMa is the fruit fly (~45 million synapses), which has 1/3300000$^{th}$ of the estimated number of human synapses. LaMa's training corpus is composed of 4.5 million distinct naturalistic images of scenes, substantially less than the plausible exposure of an adult human being if we assume that of the ~100 000 saccades we make daily[24] a large proportion will materially change the perceived image. Though neither model training nor execution are directly comparable across electronic and organic substrates, LaMa's is minimal on both counts: trained on 8 Nvidia V100 GPUs for approximately 240 hours[9], and requiring 1.7 seconds to in-paint 12 RAPM questions. Ours is therefore likely to be, if anything, an overestimate of the minimal substrate necessary to perform the task validly under test conditions.

**Discussion**

The challenge of operationalizing cognitive abilities, especially those of the most general kind, demands rigorous examination of the properties of any candidate test. Exploiting recent advances in self-supervised machine learning, here we introduce a new criterion of validity—substrate validity—determined by the correspondence between the properties of the minimal computational substrate a task requires, and those of the neural substrate plausibly engaged in the real-world exercise of the target cognitive ability. We show that arguably the best established clinical test of fluid intelligence, RAPM, is accessible *a prima vista*—'zero shot'—to a comparatively small artificial neural net, free of task-specific inductive bias, trained on only 4.5 million low-resolution images of natural scenes with a single, simple objective: natural image completion by 'in-painting' arbitrarily masked regions. LaMa's performance corresponds to a full-scale IQ of ~106, exhibits error patterns and modulation with item difficulty characteristic of human populations, and degrades with specific ablation of the capacity to integrate global spatial information. The minimal computational substrate of RAPM is thus revealed to be a machine capable of generalising the interpolation of spatial patterns from limited training on partially occluded natural scenes. This demonstrates that though RAPM may *correlate* with verbal and visuospatial reasoning[25], abstraction[26], working memory[27,28], or cognitive control[29], it does not necessarily *require* them. Equally, though RAPM may *predict* real-world aptitudes plausibly sensitive to the flexible deployment of reasoning powers, it need not necessarily *depend* on the neural systems that subserve it. We now examine thirteen aspects of the interpretation and implications of our results.

First, we should consider how a test of novel, abstract, inductive reasoning may be recast as a problem of spatial interpolation. The novelty RAPM problems introduce is of *content*, rather than *process*: the process of completing a partially occluded spatial pattern cannot be novel to anyone with reasonably rich visual experience, and the wide variety of encountered visual forms will promote invariance to content in the natural ability. Equally, the only abstraction in RAPM is the compressed representation of visual forms in terms of their edges—a low-level process in the visual system[30]—the component forms themselves may be encountered in real life. RAPM problems can be framed in terms of naturalistic, continuous images without either simplifying or complicating the organising rule: there is no reason to suppose such a formulation more or less difficult. Completing a partially obscured, spatially organized pattern from a choice of alternatives is not the same as deducing the generative rule, and need not invoke reasoning: a subject may select their answer based on what is perceived to 'fit' more or less intuitively. The criterion of correctness is simply conformance with the pattern.

Second, the minimum identified here is of the complexity of a computational substrate minimally capable of executing the task *under the appropriate criterial constraints*, not of the solution itself. A set of rules of reasoning stated in natural language may be a more compact solution than an 'in-painting' model, but the substrate of such a solution would necessarily extend beyond the rules, to the model that generates them. Only extraordinarily large foundational models such as GPT-4[31,32] are capable of such reasoning, where the complexity of the model precludes readily interpretable manipulations of capacity. Equally, we are not concerned with the well-known task of explicitly solving Raven's matrices-style problems with machines. In common with many other problems, an automated solution can here be found by embodying the relevant inductive biases into a machine[33–37], or training discriminative models on the specific task[38–40]. Though undoubtedly interesting, such models do not provide a lower bound on the complexity of a successful solution, for they constrain it either with the help of human induction or explicit training. Equally, the capacity of the minimal substrate we have identified here sets a ceiling on the minimum complexity, not the floor: it is the infimum or maximum lower bound. It is possible that even simpler models may be discovered in the future, though they cannot, of course, be crafted for the specific task.

Third, though constructed on similar principles, individual RAPM items need not be homogeneous in their solutions. In showing that LaMa can solve *some* RAPM items, we do not prove that an enhanced architecture of the same kind could solve *all*. Indeed, metacognitive analysis of human performance suggests possible heterogeneities of approach, commensurate with variations in the intricacy of the underlying rules[41]. But in the absence of a *categorial* difference it is hard to exclude the possibility that modest augmentation of LaMa style architectures could achieve substantially higher performance, while still leaving a very substantial gap in comparative substrate complexity. This ought to be an area of future study.

Fourth, that a function is not *required* by a cognitive test does not mean it is not *invoked* by it in some, or indeed all, instances. The task of finding the sum of integers from 1 to n involves—for most—maintenance in working memory of a serial chain of additions, yet—for the young Carl Gauss[42]—merely the simple arithmetic of $n(n + 1)/2$. It is possible that most subjects solve RAPM-style problems through abstract reasoning of the kind commonly assumed to be necessary, involving commensurate time and perceived intellectual effort. Equally, a test may not stress the neural substrate of a real-world aptitude but nonetheless interact with it. For example, the ability to allow the comparatively 'low-level' natural completion of a spatial pattern to guide behaviour may be kin with the cognitive flexibility challenging tasks demand, reflecting a capacity for orchestrating distinct cognitive elements. Indeed, the sense of the natural continuity of a spatial pattern, though implicit in our interactions with the visual world, is not normally surfaced as an explicit task, let alone a multiple-choice one. Such a relationship would be more robust to variation—physiological or pathological—than mere correlation but cannot otherwise be easily distinguished from it. Crucially, there is no criterion by which the mode of solution can be reliably determined. The subject's report is no guide, for the cited reason may be given *post hoc*[43], or guided by an intuitive sense of the right response: action and its explanation are not causally related[44]. Moreover, the completion of a masked pattern is simpler, more natural, and evolutionarily older than abstract reasoning, for it is implied in our ability to discern interrupted patterns of spatial organisation, such as the spatial continuity of a predator glimpsed through dense foliage, which may generate surprisingly complex 'spatial arithmetic' effects, including subtraction. In any event, where a simpler solution is both available and potentially indistinguishable from any alternative, its possibility is hard to deny or its prevalence to quantify.

Fifth, that a test may differ in its distinctive neural substrate from a real-world aptitude *does not* mean their correlation cannot be useful, both as a cognitive measure, and as an anatomical or pathological localiser. Correlations across cognitive variables are common, and it is plausible that performance on two different tasks may be similar owing not to a shared neural substrate, but to similar constitutional or developmental shaping of distinct substrates. The clinical utility of a cognitive test is primarily determined by its psychometric and localizing properties: if RAPM is the best *available* guide to (say) the probability of return to gainful employment or the lateralisation of a focal pathological process, it is no less so now that its substrate has been shown to be radically simpler than previously thought. But the fact demonstrates a better test *could* be possible, for the closer a test cleaves to the critical substrate of an ability the more accurate it should be. Moreover, relying on mere correlation is hazardous in a clinical context, where anatomically specific patterns of pathological disruption need not respect commonalities of constitutional or developmental origin. Patients with normal RAPM performance in the context of profound wider cognitive dysfunction provide a vivid illustration of this problem[45].

Sixth, that a task may critically depend on the interpolation of spatial patterns does not imply it can be dismissed as merely 'visuospatial' or assumed to depend on presumptively 'low-level' neural substrates designated as 'visuospatial' on other empirical grounds. Constraining a spatial interpolant on a set of presented possibilities is a kind of categorial counterfactual *imagination* (*not* mental imaging): i.e. 'what would the coherence of a pattern be given completion X vs Y?' If this is visuospatial then—computationally simple or complex—it is quite unlike the vast majority of visuospatial tasks in the literature and cannot be grouped with them, in either behavioural or neural terms. Dependence on anterior vs posterior cortical areas, for example, does not become less or more plausible on that account, no reliable mapping of computational to neural substrates is possible here outside the lesion-deficit study of the test itself or others specifically like it[46].

Seventh, a neural implementation of the minimum computational substrate would be expected to be an order of magnitude faster than commonly observed human response times[47,48], permitting uncharacteristically rapid, implicit, intuitive solutions. Exploration of the effectiveness of explicitly prompting such a strategy is an interesting empirical question; its rarity does not necessarily imply infelicity, as the Gaussian summation example shows[42]. Indeed, it may be invoked where a pathological deficit bars the explicit reasoning approach most respondents will naturally assume to be expected of them when presented with an 'intelligence test'. A further empirical question to be addressed in future studies is the relationship between the correctness of an answer and the ability to state the underlying rule. The prevalence of discrepancies between the two could provide an index of the prevalence of intuitive solution strategies.

Eighth, we should consider the evidence that our LaMa evaluation is a fair test of RAPM's substrate validity. LaMa's fundamental architecture—a fully-convolutional feed-forward net—demonstrably contains no inductive bias intelligible as abstract reasoning, on geometric figures or any other kind. It is given no opportunity to learn RAPMs or any other putative reasoning problems before the test—no test-specific training is performed, and each item is evaluated independently—and the corpus on which it learns its only task—'in-painting'—is composed entirely of low-resolution images of natural scenes. The number of distinct training images, 4.5 million, constitutes a substantially poorer visual exposure than an adult human being is likely to have received, as is their variety and scope. If the average feature film has ~1000 shots and ~100 000 frames, the number of distinct images should be at least ~10 000, yielding the equivalent of viewing only ~500 films *as the totality of visual experience*[49]. The training process, and the objective function it optimizes, are confined to the retrieval of a perceptually coherent natural image from an arbitrarily masked version: none involves either symbolic abstraction or any simulacrum of reasoning. Since LaMa cannot receive instructions or provide answers, its solutions must be inferred from image-based comparison between the available choice of answers and the 'in-painted' image. But this makes the task much harder, for the model has to 'draw' the response rather than merely choose from a set of options. This means the baseline odds of obtaining a correct answer are substantially lower than the 0.125 guess rate 8 possible answers implies, for the model must produce a coherent image first. It also provides a qualitative indication of the model's belief, illustrated in Figure 3, independently of the quantitative metrics. The permitted time and computational resource are also modest in comparison with human context. In short, no aspect of the model or its deployment provides any advantage to the machine: quite the opposite.

Ninth, we should examine the *quality* of LaMa's behaviour in comparison with normal and impaired human agents. LaMa exhibits a clear monotonic relation to item difficulty; the threshold of the psychometric function falls within the 95% Bayesian confidence interval of healthy human performance. Moreover, the erroneous choices LaMa makes under degradation of its architecture significantly correlate with those made by patients with right frontal damage, the area recently shown to be most strongly associated with reduced RAPM performance[5]. As Figure 3 shows, where LaMa offers an answer—either correct or incorrect—'in-painted' regions are not random, but typically correspond to one of the choices on offer, mirroring a human subject's plausible belief. There are no readily identifiable grounds on which an external observer could judge LaMa to differ from a human requested to provide an answer in drawn form.

Tenth, consistent with the proposition that RAPM minimally requires the integration of global spatial patterns, the form of degradation that impairs LaMa's RAPM performance the most—while retaining its fundamental capacity to 'in-paint'—is ablation of its Fast Fourier Convolutional modules, which implement a perceptual organisation with wider receptive fields, and facilitate sensitivity to global repetitive patterns. Crucially, it is precisely this form of degradation that replicates the error patterns of patients with right frontal damage[5]. Moreover, the qualities of these errors characteristically show extrapolation of local patterns, with failure to incorporate more widely distributed spatial structure (Figure 3). This coheres with known right frontal lateralization of global vs local spatial processing[50], and spatial working memory, navigation, and complex visuospatial processing[51–56]. Indeed, interference from undefined global effects on a simple, purely locally-driven, automatic visuomotor action is exquisitely sensitive to right frontal white matter microarchitecture[57]. Though intriguing, the radical architectural differences between models such as LaMa, and scales of neural organisation where sensitivity to complex spatial frequencies is observed, are too great to allow meaningful parallels to be drawn mechanistically rather than merely functionally[58]. Sensitivity to global repetitive patterns may be achieved computationally by means other than FFC; equally, both FFC and other forms of signal decomposition may be implemented by a neural network[59].

Eleventh, we introduce the use of what might be termed 'transfer lesion-deficit mapping', to investigate the human neural dependents of behavioural patterns exhibited by a machine substrate. The approach does not presuppose or infer mechanistic similarity between neural and machine substrates, but rather provides a means of generating anatomically framed hypotheses about the neural dependents of a test shown by computational substrate analysis to be open to a specific kind of solution. For example, where computational analysis reveals a heterogeneity of possible solutions, a corresponding heterogeneity of neural dependents may indicate a corresponding heterogeneity of the critical substrate that need not be evident from human behavioural patterns alone. Though our lesion-deficit analysis here suggests a correspondence between LaMa errors and right frontal damage, no strong conclusions can be drawn owing to the modest size of the positive class in the cohort.

Twelfth, we formalise the criteria a licit evaluation of the substrate validity of a test must satisfy. The bare fact of an automated computational solution is obviously not sufficient. Since our interest is typically in the lower bound on the computational nature or complexity of a test—the minimal substrate—we need tight restrictions on the validation process, the characteristics of the model, and the interpretation of the results. It should be plain that the candidate validation substrate must be exposed to data inputs, in development and testing, of the same kind as, and no greater number than, a human subject, i.e. there should be **data parity**. The performance of any model may otherwise represent superhuman over-training of limited translatability to the neural domain. It should also be plain that the model cannot be equipped with task-specific inductive biases, for then the solution is incorporated into the design and the effective substrate extends to the mind of the designer, i.e. there should be **inductive bias parity**. A model may otherwise simply implement a specific human-derived rule, as an electronic calculator embodies arithmetic. Especially where the task is conceived to be novel, the model must not be explicitly trained on test questions and answers or anything artificially resembling them, i.e. training must be under pure **self-supervision**. A model may otherwise overfit to a solution implicitly provided by the human designer in training. The output of the model when performing the test task must be at least as expressive as a human's, so that the scoring itself does not inject inductive bias into the process, i.e. there must be **target parity**. Our estimates of the performance of a substrate may otherwise be inflated by the human intelligence required to interpret the answers. Finally, the conditions of the test, such as available time, should be comparable, i.e. there should be **test parity.**

Finally, we should stress that an evaluation of substrate validity concerns the test, not the cognitive powers the test purports to measure. That a computational substrate comparable to the fruit fly's nervous system can complete a test of fluid intelligence neither renders flies more intelligent nor implies flexible problem-solving loads only 1/3300000$^{th}$ of the brain's capacity. Nor does it prove that the substrate itself has acquired the cognitive powers, in mysteriously emergent form: to conclude so is to commit the familiar fallacy of affirming the consequent[60]. Nonetheless, the performance of any model architecture, especially under ablation, may cast further light on the nature of the problems our neural equipment has evolved to solve, enabling us to focus on the essential neural capacities.

The study of human intelligence has historically catalysed the development of the artificial kind. The relationship has become bi-directional now that advances in technology allow us to create computational substrates with the capacity to respond to cognitive instruments previously only applicable to human beings. Instead of assuming the cognitive dependents of a test, we can directly quantify them, through substrate validation implemented with flexible self-supervised neural network architectures. The approach is naturally extensible to generating novel tests whose computational burden can be explicitly modelled, and assured to stress a range of powers that is both defined and fully inclusive. As our engineering of machines brings them closer to human beings, each will increasingly cast reciprocal light on the other, illuminating our understanding of cognition, both human and artificial.

**Methods**

*Participants*

Data from 89 patients with focal right frontal lobe lesions who attended the Neuropsychology Department of the National Hospital for Neurology and Neurosurgery, London, UK, was retrospectively obtained subject to previously reported inclusion and exclusion criteria[5]. A set of 532 healthy control participants was recruited to provide normative RAPM data for a culturally and ethnically diverse population across the full adult lifespan, as reported in a separate study[7]. This group was filtered to include only those patients for whom item-wise RAPM scores were available, resulting in 58 right frontal lesion patients and 315 healthy controls. To ensure that statistical effects in the linkage of model errors to human participant characteristics were evaluated robustly and not simply the consequence of comparing any lesion with healthy controls, a further group of patients with a focal left frontal or posterior lesion was included, again retrospectively obtained subject to previously reported inclusion and exclusion data[5]. For all participants, data on age; gender; number of years spent in education; and NART IQ, a measure of premorbid level of function, was extracted in addition to RAPM scores. For the purposes of error analysis only, to ensure that any observed effects were not simply due to the presence of any lesion vs healthy controls, a further set of patients with focal lesions in left frontal or posterior areas were added for statistical comparison; these were again previously reported[5]. Participants who did not have completely recorded error data were excluded (healthy control = 244, right frontal lesion = 49, left frontal or posterior lesion = 85).

The study was approved by The National Hospital for Neurology and Neurosurgery and Institute of Neurology Joint Research Ethics Committee and conducted in accordance with the 'Declaration of Helsinki'.

*Behavioural investigations*

RAPM is a neuropsychological test administered to quantify education- and culture-independent fluid intelligence in people of expected higher than average intelligence, or who have achieved high intelligence scores on Raven's Standard Progressive Matrices (RSPM)[61]. In the time-constrained clinical context, RAPM Set 1 (/12) is validated as a useful estimate of the longer form test[6,7]. The test consists of a 3x3 pattern of abstract shapes; participants are asked to choose one of eight possible shapes to complete the pattern, a task proposed to require multiple steps of abstract reasoning.

*'In-painting' model testing on RAPM*

RAPM questions were cropped to 512x512 pixels and a rectangular mask was drawn over the area to be completed: the 9th cell of the 3x3 image matrix. Images and masks were tested according to the procedure described in https://github.com/saic-mdal/lama for the most performant LaMa 'in-painting' model (51M parameters), which had been trained exclusively on the Places dataset[62], a large collection of images across 400 built and natural scene categories, from within buildings, to transport networks, to landscapes. The resulting in-painted images were registered to matrices completed by each of the 8 response options, using ORB matching[63] by Hamming Distance and warping by RANSAC homography[64]. A panel of 5 metrics were then calculated between registered in-painted images and response images cropped to the 9th cell answer region: Hausdorff distance (HD)[65]; mean squared error (MSE); Wasserstein distance (WD)[66]; Erreur Relative Globale Adimensionnelle de Synthèse (ERGAS)[67]; and normalised mutual information (NMI), chosen to provide a broad set of image comparisons. The response option corresponding to the smallest value of HD, MSE, WD, and ERGAS, and the largest value of NMI, was selected, and the mode response was selected as the model 'answer'.

To understand the contribution of model components to RAPM performance, the above procedure was repeated for a model ablated of FFC (74M parameters), which is reported to enable learning of periodic information across the entire receptive field at all model layers; and also a model of smaller parameter size, batch size and training set size, which retained the FFC (27M parameters).

*Psychometric function analysis*

Group-level psychometric functions were obtained for healthy controls and right frontal lesion patients with respect to the order of item difficulty observed for the former. For each group, posterior distributions of the threshold were derived from Bayesian models of the underlying psychometric function of logistic form with four parameters—guess rate (fixed at 0.125), threshold, slope, and lapse rate—implemented in Psignifit v3.0 for MATLAB[68]. We report 95%

Bayesian credibility intervals at a threshold of 0.5 probability of success. To enable psychometric analysis of the LaMa model, which has a non-stochastic output, we applied random Gaussian noise or randomly altered brightness 50 times per question, and scored the closest answer by taking the mode of a panel of metrics as previously described. The number of incorrect scores /50 were then provided as input to the psychometric function generating procedure as per the human participant analysis.

*Analysis of errors*

Greater insight into performance may be obtained by analysis of the types of incorrect responses selected. We sought to understand the distribution of error patterns by plotting heatmaps of each 1-of-8 option selection frequency in different groups. First, questions were ordered by normative difficulty—the number of healthy controls that answered the question correctly—and second, response options were ordered by selection frequency in the healthy control group. Questions and responses were plotted as heatmaps by frequency for the healthy control group; the right frontal lesions were plotted as heatmaps according to the positions determined by the healthy control group. Statistically significant differences in response frequency were investigated by applying z-test for proportions for the 12x8 grid of response options, and the alpha value of 0.05 was adjusted for multiple comparisons using Benjamini/Yekutieli false discovery rate correction. Note this 12x8 errors analysis differed from previously reported errors analysis on this cohort, in which items were grouped into 3 difficulty-stratified groups[5].

In a corresponding analysis, where LaMa, LaMa FFC-ablated, and LaMa small made a clearly visible 'in-painting' attempt, the in-painted answer was scored by a panel of distance metrics as previously described. Note that copyright restrictions preclude the publication of actual RAPM questions or attempted answers, whether correct or not, except for the first, introductory, question shown here. Human participants (all healthy controls and all patients with frontal or posterior lesions) were grouped into those who made any of the same errors as each of the LaMa models, and those who did not. Differences in the group characteristics–age, education years, NART IQ, sex, and presence of right frontal lesion–were evaluated for significance using chi-squared test for categorical variables, and Mann Whitney U test for continuous variables, Benjamini/Yekutieli false discovery rate correction for 0.05 alpha.

*Neuroimaging investigations*

The lesion maps of human participants who made errors identical with the FFC-ablated model were further investigated with graph lesion-deficit modelling, a method described and validated in detail elsewhere[5] that employs Bayesian Stochastic Block Models[14,69,70] to infer distinct patterns of connectivity based on graph community detection. Modelling correlations between voxels within a graph permits the removal of spatially confounding effects arising from pathologically driven patterns of collateral damage: a problem that has been repeatedly shown to corrupt lesion-deficit maps derived with mass-univariate methods such as VLSM.

Imaging data was available for 108 patients; 14 of whom made FFC-ablated errors, and 94 of whom did not. Lesions were manually segmented using MIPAV (https://mipav.cit.nih.gov/), reviewed by a neurologist (PN), non-linearly normalized to Montreal Neurological Institute (MNI) stereotaxic space at 4×4×4 mm resolution using SPM-12 software (http://www.fil.ion.ucl.ac.uk), and transformed into a voxel-wise adjacency matrix. An undirected hypergraph was then constructed with voxels as nodes, and two edge covariates—the probability of two voxels being co-lesioned in general, and the probability of two voxels being co-lesioned with respect to the cognitive deficit (here a FFC-ablated type error)—distributed in separate layers of a Layered Stochastic Block Model[70] so as to distinguish between pathology- and deficit-specific anatomical effects[71]. After an initial fit, the model was optimised by simulated annealing using default inverse temperature of 1 to $10^6$, and a wait criterion of 100 iterations past a record-breaking event to ensure equilibration was driven by changes in entropy[72]. Bayesian model comparison was used to determine the superiority of this model over a null where lesion co-localisation and cognitive deficit edge layer connections are randomised[72]. The layered model exhibited substantially lower entropy than the layer-randomised null (1220000 vs 1280000). To visualise the inferred communities, we backprojected the incident edge weights onto the brain, deriving the mean and 95% credible intervals for comparison.

*Analytic Environment*

Psychometric function modelling was conducted with Psignifit v3.0 in MATLAB (https://psignifit.sourceforge.net/); all other analyses were written in Python 3.9. Data processing was performed using Pandas[73], NumPy[74], image processing and analysis with OpenCV-Python[75], MONAI[76], sewar, and Scikit-image[77]; graph modelling with graph-tool[78]; statistical testing with SciPy[79] and statsmodels[80]; and visualisation using Matplotlib[81], and Seaborn[82]. The hardware specification used was as follows: 96 Gb RAM, Intel Xeon E5-2620 CPU at 2.10 GHz, and Nvidia GeForce GTX 1080.